\documentclass[runningheads]{llncs}
\usepackage{graphicx}
\usepackage[export]{adjustbox}
\usepackage{comment}
\usepackage{subcaption}
\usepackage{amsmath,amssymb} 
\usepackage{color}
\usepackage{bbm}

\newcommand{\floor}[1]{\left\lfloor #1 \right\rfloor}

\usepackage[width=122mm,left=12mm,paperwidth=146mm,height=193mm,top=12mm,paperheight=217mm]{geometry}

\DeclareMathOperator*{\argmax}{arg\,max}

\newcommand{\algName}{{NoiseRank }}

\begin{document}
\pagestyle{headings}
\mainmatter

\title{NoiseRank: Unsupervised Label Noise Reduction with Dependence Models} 


\author{
Karishma Sharma \\
University of Southern California\\
{\tt\small krsharma@usc.edu}
\and
Pinar Donmez \\
Facebook AI\\
{\tt\small pinared@fb.com}
\and
Enming Luo \\
Facebook AI \\
{\tt\small eluo@fb.com}
\and
Yan Liu \\
University of Southern California\\
{\tt\small yanliu.cs@usc.com}
\and
I. Zeki Yalniz \\
Facebook AI\\
{\tt\small izy@fb.com}}

\institute{Paper ID \ECCVSubNumber}

\author{Karishma Sharma\inst{1} \and
Pinar Donmez\inst{2} \and
Enming Luo\inst{2}\and
Yan Liu\inst{1}\and
I. Zeki Yalniz\inst{2}}\institute{University of Southern California, USA \and
Facebook AI\\
\email{krsharma@usc.edu,pinared@fb.com,eluo@fb.com,yanliu.cs@usc.edu,izy@fb.com}\\
}
\maketitle

\begin{abstract}
Label noise is increasingly prevalent in datasets acquired from noisy channels. Existing approaches that detect and remove label noise generally rely on some form of supervision, which is not scalable and error-prone. In this paper, we propose NoiseRank, for unsupervised label noise reduction using Markov Random Fields (MRF). We construct a dependence model to estimate the posterior probability of an instance being incorrectly labeled given the dataset, and rank instances based on their estimated probabilities. Our method 1) Does not require supervision from ground-truth labels, or priors on label or noise distribution. 2) It is interpretable by design, enabling transparency in label noise removal. 3) It is agnostic to classifier architecture/optimization framework and content modality. These advantages enable wide applicability in real noise settings, unlike prior works constrained by one or more conditions. NoiseRank improves state-of-the-art classification on Food101-N ($\sim$20\% noise), and is effective on high noise Clothing-1M ($\sim$40\% noise).


\end{abstract}

\vspace{-5mm}

\section{Introduction}\textbf{}
\vspace{-5mm}

\begin{figure}[t]
\begin{center}
\includegraphics[width=0.8\linewidth,height=6cm]{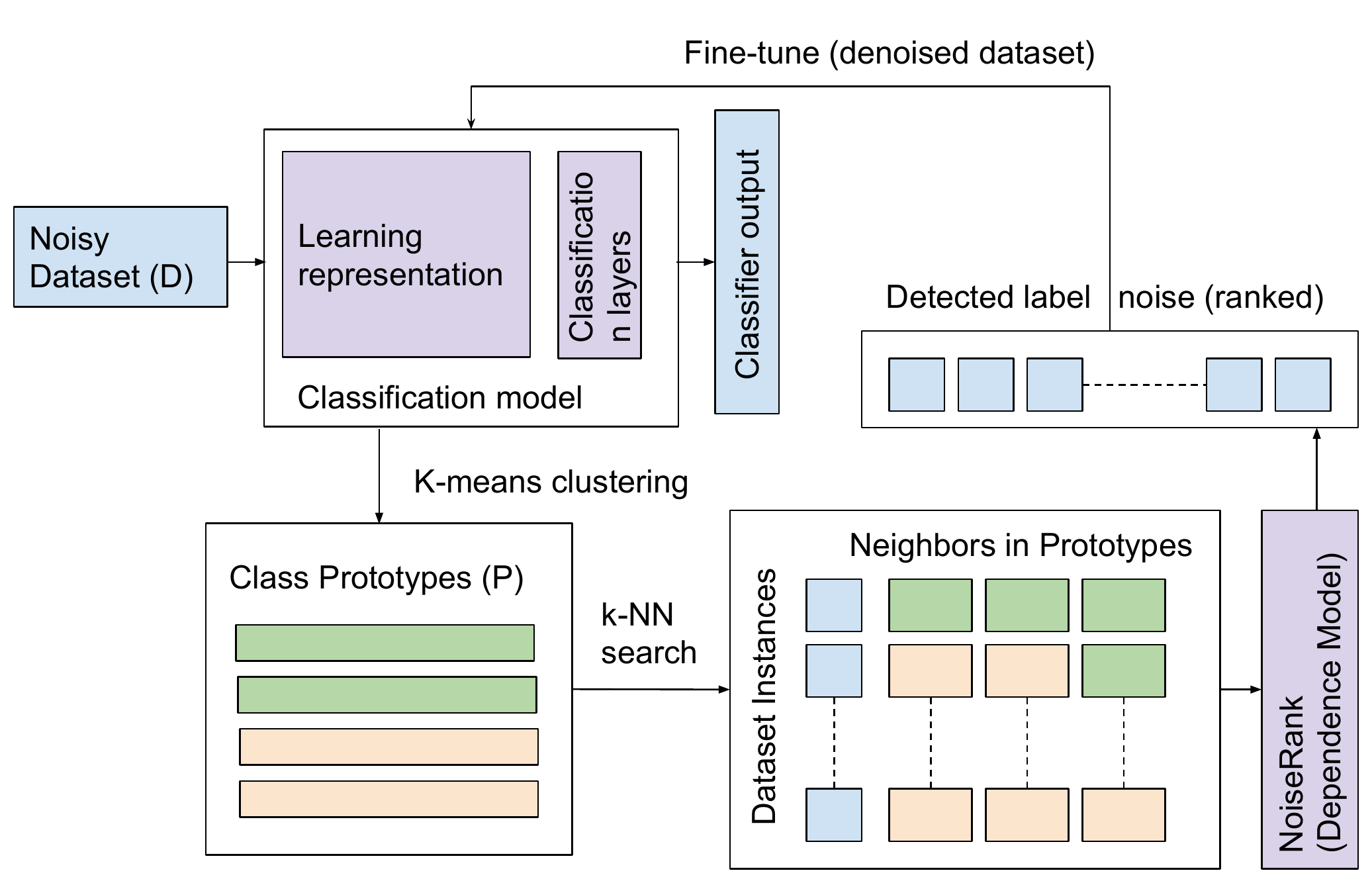}
\end{center}
   \caption{Illustration of ``NoiseRank'' framework for unsupervised label noise reduction and iterative model training
   (interpretable, and agnostic to classification model architecture and optimization framework)}
   \label{fig:iterative_labelnet}
\end{figure}

Machine learning has become an indispensable component of most applications across numerous domains, ranging from vision, language and speech to graphs and other relational data \cite{pouyanfar2019survey}. It has also led to an increase in the amount of training data required to effectively solve target problems. Labeled datasets, typically obtained through manual efforts, are prone to labeling errors arising from annotator biases, incompetence, lack of attention, or ill-formed and insufficient labeling guidelines. The likelihood of human errors increases in domains with high ambiguity \cite{ross2017measuring,waseem2016you}. Additionally, there is an increasing dependence on automated data collection such as employing web-scraping, crowd-sourcing and machine-generated labeling \cite{corbiere2017leveraging,vondrick2013efficiently}. However, the cheap but noisy channels have made it imperative to deal with incorrectly labeled samples.

Existing literature either focus on training noise-robust classifiers \cite{arazo2019unsupervised,guo2018curriculumnet,jiang2018mentornet,jindal2019effective,li2019learning}, or attempt to reduce or correct label noise in the dataset generally with some form of supervision \cite{lee2018cleannet,patrini2017making,veit2017learning,xiao2015learning}. However, attention is shifting towards \textbf{unsupervised label noise reduction} due to obvious practical benefits. Most earlier methods use some form of supervision, either from verified or clean labels, or priors on the label/noise distribution, in order to guide the detection of mislabeled examples. In this work, we propose a fully unsupervised approach for label noise detection using Markov Random Fields (MRF), also known as dependence models, which provide a generic framework for modeling the joint distribution of a large set of random variables. We formulate a dependence model 
to estimate the posterior probability of an instance being incorrectly labeled,  given the dataset, and rank instances based  on  the  estimated  posterior.
We provide an iterative framework for label noise reduction using our dependence model for noise ranking, as shown in Fig.~\ref{fig:iterative_labelnet}. The iterative framework is used to first learn instance representations from the noisy dataset, and detect label noise, then fine-tune on denoised (cleaned) subsets in order to improve classification and learned representations, which iteratively improve label noise detection.

Our approach addresses several shortcomings of existing methods. First, our proposed method ``NoiseRank" removes dependence on supervision for label noise detection. This allows wider practical applicability of our method to real domains. In contrast, most supervised approaches dealing with label noise are error-prone and hardly scalable. Second, our proposed framework for label noise ranking and improving classification is agnostic to both the classifier architecture and its training procedure. The implication of this is that we can train classifiers on any domain (image, text, multi-modal, etc.) within the same framework, using any standard classification architecture and optimization framework. In comparison, methods such as \cite{arazo2019unsupervised,guo2018curriculumnet,li2019learning,jindal2019effective} require careful network initialization and regularization of the loss function for optimization. Lastly, NoiseRank's underlying algorithm and its output are human interpretable by design.
Our main contributions are summarized as follows:
\begin{itemize}
    \itemsep0em
    \item A fully unsupervised label noise detection approach which is a probabilistic dependence model estimating the likelihood of being mislabeled. It does not require ground-truth labels, or priors on label or noise distribution. 
    \item The proposed framework is generic, i.e., independent of application domain and content modality, and applicable with any standard classifier model architecture and optimization framework, unlike many recent unsupervised approaches \cite{yi2019prob,han2019deep,arazo2019unsupervised,guo2018curriculumnet,tanaka2018joint}. 
    \item Its underlying algorithm and output are human interpretable. Again many unsupervised methods do not incorporate interpretability \cite{yi2019prob,tanaka2018joint,jiang2018mentornet}, which reduces their transparency in label noise detection and ranking.
    
    
    
    \item Experiments on real noise benchmark datasets, Food101-N ($\sim$20\% noise) and Clothing-1M ($\sim$40\% noise) for label noise detection and classification tasks, which improved state-of-the-art classification on Food101-N.

\end{itemize}

\section{Related Work}

\textbf{Robust and noise-tolerant classifiers:} Methods that focus on training noise tolerant or robust classifiers attempt to directly modify the training framework for learning in the presence of label noise. \cite{jindal2019effective} introduces a non-linear noise modeling layer in a text classifier architecture to encode the distribution of label noise. \cite{arazo2019unsupervised} fits a beta mixture model on the training loss distribution to estimate the likelihood of label noise and uses that to guide the classifier training with a carefully selected loss function based on bootstrapping \cite{reed2014training} and mix-up data augmentation \cite{zhang2017mixup}. Approaches based on meta-learning and curriculum learning are also studied for modifying the training procedure, where training samples are either ordered based on learning difficulty or mixed with synthetic noise distributions \cite{guo2018curriculumnet,li2019learning,jiang2018mentornet}. However, these methods limit the choice of classifier architectures, and furthermore are known to work only with careful initialization \cite{jindal2019effective} and regularization \cite{arazo2019unsupervised} needed for convergence.

\textbf{Label noise reduction/correction:} Other methods, including ours, are based on label noise reduction, that attempt to detect, and remove or correct label noise. Prominent approaches utilize supervision to guide label noise detection. \cite{veit2017learning,xiao2015learning} require clean (ground-truth) labels for a subset of the data to learn a mapping from noisy to clean labels. \cite{lee2018cleannet} requires binary verification labels instead, which indicate whether the given label is correct or noisy, in order to train an attention mechanism that can select reference images as class prototypes, and learn to predict if a given label is noisy. Similar to \cite{lee2018cleannet}, \cite{han2019deep} uses prototypes (more than one per class) to generate corrected labels which are then employed to iteratively train a network. However, \cite{han2019deep} does not rely on any supervision or assumptions on the label distribution. As compared to \cite{han2019deep}, our method uses standard cross-entropy loss, whereas their framework is based on self-supervised learning, limiting flexibility on classifier optimization framework. 

Another iterative approach is of \cite{yi2019prob}, which updates both network parameters and label distributions to iteratively correct the noisy labels. \cite{patrini2017making} relies on the availability of a noise transition matrix for loss correction when training a classifier, which specifies the noise distribution in terms of the probability of one class being mislabeled as another. \cite{xia2019anchor} employs a deep learning based risk consistent estimator to fine-tune a noise transition matrix.
One type of unsupervised approach is outlier removal \cite{xia2015learning,platt1999estimating}. However, outliers are not necessarily mislabeled and removing them presents a challenge \cite{frenay2013classification}. There are also several methods addressing instance selection for kNN classification, which retain a subset of instances that allow correct classification of the remaining instances \cite{hart1968condensed,gates1972reduced,delany2004analysis,franco2010data,nanni2011reduced}, or remove instances whose labels are different from the majority labels of their nearest neighbors \cite{wilson1972asymptotic,muhlenbach2004identifying,lallich2002improving}. However, the proposed heuristics have been criticized for removing too many instances or keeping mislabeled instances \cite{frenay2013classification}. Our approach is related to these methods but focuses on leveraging both label (in)consistencies to globally rank noisy candidates and more effectively detect mislabeled instances even without any supervision. Weakly-supervised methods based on classification filtering such as \cite{thongkam2008support} remove samples misclassified by SVM trained on the noisy data. However, it could amount to removing non-noisy hard samples or not removing noisy samples that the classifier mistakenly fits.

\section{Unsupervised Label Noise Reduction and Model Training Framework}
Our ultimate goal is to learn an effective classifier from the noisy labeled dataset without any form of human supervision (i.e., label verification), prior knowledge on the target domain, or label/noise distribution. In this section, we elaborate our multi-step model training framework. As is illustrated in Fig. \ref{fig:iterative_labelnet}, we first describe vector representation, which is necessary for similarity measure and label prediction in our framework. Next, our proposed probabilistic dependence model ``NoiseRank'' is elaborated for ranking dataset examples based on their likelihood of being noisy. Finally, we discuss the iterative model training steps.

\subsection{Vector Representation} \label{sec:rep_learning}

The vector space representation (i.e., embeddings) is a core component of our design because we rely on the vector representations to determine content similarity between examples in the given noisy dataset. Our framework is agnostic to any modality as well as the solution for learning representations. However, a high-quality representation improves the similarity measure and thus our unsupervised method for label noise detection. In Sec. \ref{sec:labelnet_classification}, we will discuss how we improve the representation through iterative training.

With the vector representation, we could determine the content similarity. More formally, let the noisy labeled dataset be denoted as $\mathcal{D} = \{(x_i, y_i)\}_{i=1}^{N}$ where $x_i\in \mathbb{R}^m$ is the vector representation for example $i$, and $y_i \in \{1,2 \dots C\}$ is the given label (potentially incorrect) with $C\geq2$ being the total number of classes in the dataset. In this work, we define instance similarity in terms of Euclidean distance between $x_i$ and $x_j$ as $d(x_i, x_j) = \| x_i - x_j \|_2$.
\subsection {Label Noise Detection}


In this section, we first describe our process for generating class prototypes, that are representative instances selected for each of the $C$ classes in the dataset; followed by our non-parametric approach to generate label predictions, $y_i^\prime$, for each prototype $i$. Let $Y = \{y'_i\}_{i=1}^P$ denote the predictions. Next, we elaborate the proposed dependence model, named NoiseRank, to globally rank the dataset examples based on their likelihood of having incorrect labels given their vector representations, labels and predictions. 



\subsubsection {Generating class prototypes} 
\label{sec:class_prototypes} Each of the $C$ classes in the dataset can be represented by a set of class prototypes, i.e. a representative subset of instances in that class. We select the prototypes using K-means clustering on the vector space representations of instances in each class, given by the noisy labels. As a rule of thumb \cite{kodinariya2013review}, we select $\floor{\sqrt{\rho/2}}$ cluster centroids per class, where $\rho$ is the average number of instances per class in the dataset. Selecting class prototypes is beneficial towards improving scalability when the number of dataset instances grows.
We find that it is also important for robustness in high noise datasets, and K-means based selection is effective compared to randomly selected prototypes.

\subsubsection {Generating label predictions} 
\label{sec:label_pred} For each prototype instance $i$ represented by vector $x_i$, we generate the predicted label  $y_i^\prime$ by a weighted k nearest neighbor classifier, as specified in Eq. \ref{weighted_knn}.

\begin{equation}\label{weighted_knn}
    y_i^\prime = \argmax_{v \in \{1,2\dots C\}} \sum_{x_j \in \mathcal{N}(x_i)} \kappa(x_i, x_j) \mathbbm{1}\{y_j=v\}
\end{equation}

where $\mathbbm{1}$ is the indicator function, and the distance kernel function $\kappa(x_i, x_j)$ is used to weigh the contribution of each neighbor $x_j$ in the neighborhood $\mathcal{N}$ comprising the $k$ nearest neighbors of $x_i$:  

\begin{equation} \label{eq:dist_kernel}
    \kappa(x_i, x_j) = \frac{1}{b + d(x_i, x_j)^e}
\end{equation}
where $d(x_i, x_j)$ is the distance function discussed in Sec.~\ref{sec:rep_learning}, and $b>0$ and $e>0$ are parameters for the bias and weight exponent, respectively. The kernel function is negatively correlated to the distance function. For example, when $e=2$, the kernel will be inversely proportional to the squared distance between the instances. Since $0 \leq d(x_i, x_j) \leq \infty$, by setting a positive bias $b$, we can prevent $\kappa(x_i, x_j)$ from being undefined when $d=0$.

\subsubsection {Dependence Model Formulation} \label{sec:dep_formulate}

Our formulation is to estimate the posterior probability $P(x_{i}, y_{i}|\mathcal{D}, Y)$ that indicates the likelihood of label noise for all examples $(x_{i},y_{i})$ in the dataset $\mathcal{D}$ and rank them based on this estimate. 
For this purpose, we use Markov Random Fields (also known as MRFs or ``dependence models'' \cite{MetzlerMRF05,YalnizPAMI19}) which provide a generic framework for modeling the joint distribution of a large set of random variables. 

In dependence models, conditional dependencies are defined only for certain groups of random variables called ``cliques'', and are represented with edges in an undirected graph. We represent the graph with $G$ and the cliques in the graph as $C(G)$ in our formulations. For each type of clique $c\in C(G)$, we define a non-negative potential function $\phi(c;\Lambda)$ parameterized by $\Lambda$. The joint probabilities are estimated based on the Markov assumption as follows:

\begin{equation} \label{eq:mrf}
P(x_{i},y_{i},\mathcal{D},Y) = \frac{1}{Z} \prod_{c\in C(G)} \phi(c;\Lambda)
\end{equation}
where $Z = \sum_{x_{i},y_{i},\mathcal{D}, Y}\prod_{c\in C(G)} \phi(c;\Lambda)$ is a normalization term.
Computing $Z$ is very expensive due to the large number of summands. Since our aim is to rank examples in the dataset based on their posterior probabilities $P(x_{i},y_{i}|\mathcal{D},Y)$ and ignoring $Z$ in this formulation does not change the ranking result, the posterior probability is estimated as follows:  
\begin{eqnarray}
P(x_{i},y_{i}|\mathcal{D},Y) &=& \frac{P(x_{i},y_{i},\mathcal{D},Y)}{P(\mathcal{D},Y)} \\ \nonumber
 & \; {\buildrel {rank} \over =}\; & \log P(x_{i},y_{i},\mathcal{D},Y) - \log P(\mathcal{D},Y) \\
 \nonumber
 & \; {\buildrel {rank} \over =}\;& \sum_{c\in C(G)} \log \phi(c;\Lambda)
\end{eqnarray}
where $\; {\buildrel {rank} \over =}\;$ indicates rank equivalence. The formulation is a sum of logarithm of potential functions over all cliques. For simplification purposes, the potential function is assumed to be $\phi(c;\Lambda) = \exp(\lambda_c f(c))$,
where $f(c)$ is the feature function over the clique $c$ and
$\lambda_c$ is the weight for the feature function.
The final ranking function is computationally tractable and linear over feature functions:
\begin{equation}
P(x_{i},y_{i}|\mathcal{D},Y) \; {\buildrel {rank} \over =}\; \sum_{c\in C(G)}
\lambda_c f(c) 
\label{eq:rankfunction}
\end{equation}

Depending on the choice of the feature functions and their corresponding weights, the final ranking score in~\ref{eq:rankfunction} can be negative. In the next subsection, we elaborate our dependence model for the task of label noise detection by explicitly defining each clique, its feature function $f(c)$ and the corresponding weight $\lambda_c$.




\subsubsection{Dependence Graph Construction}

\begin{figure}[t]
\centering
\includegraphics[width=85mm]{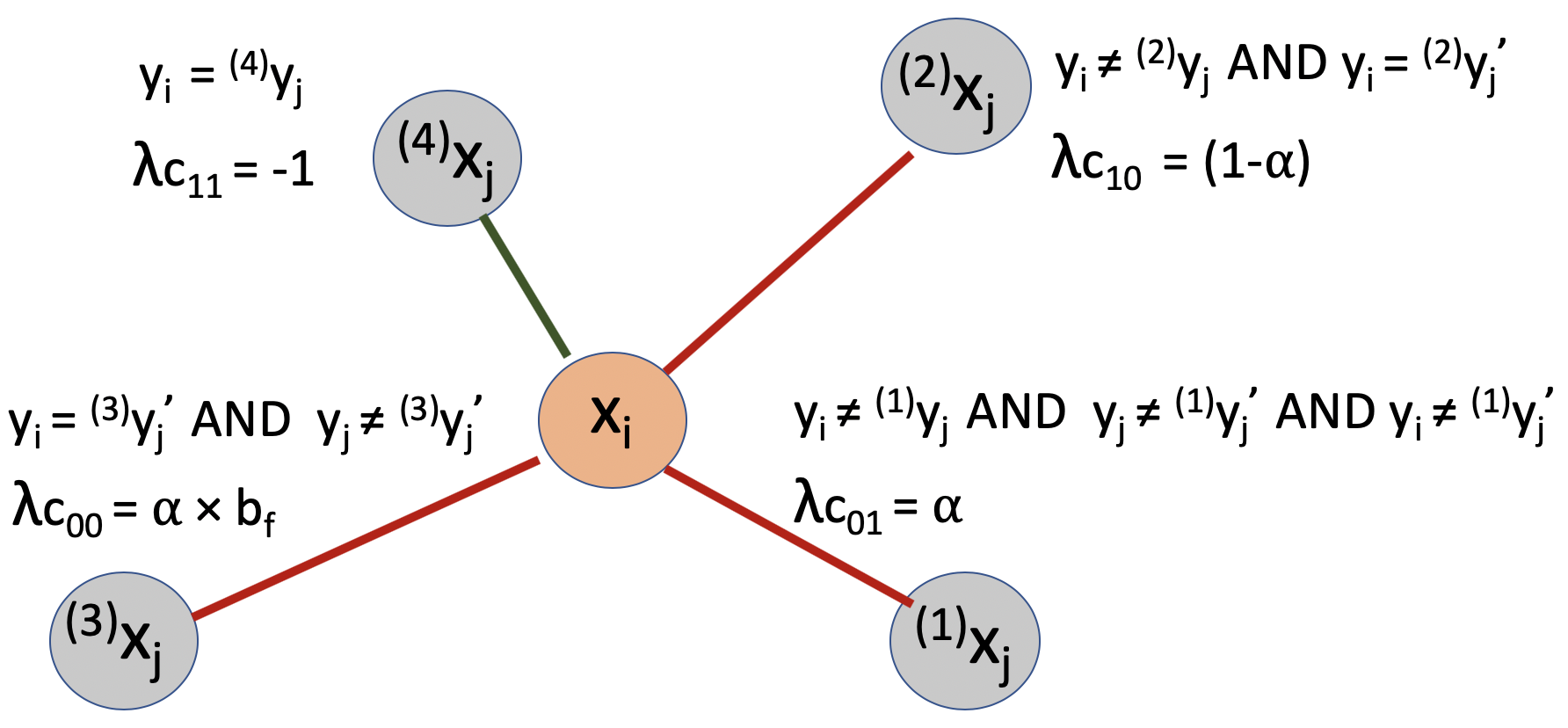}
\caption{The dependence graph illustrated for a given example $i$ in a database containing five examples. Clique types and weights are determined based on the given $y_i$ and $y_j$ and predicted labels $y_j^\prime$. Edge lengths indicate distance in the vector representation space}
\label{fig:dependence_graph}
\end{figure}




In our formulation, we define cliques between all pairs of examples $(i,j)$ where $i \neq j$ and $i \in \mathcal{D}, j \in P$ for dataset $\mathcal{D}$ with size $N$ and set of all prototypes $P \subseteq \mathcal{D}$. There are $\mathcal{O}(N|P|)$ cliques defined in the dependence graph. 
Each example is associated with its given label $y_i$ and each prototype is associated with its given label $y_j$ and predicted label $y_j^\prime$, which are used for determining the clique weights as shown in Fig. \ref{fig:dependence_graph} and explained below.
All cliques are assumed to share the same feature function $f(c) = \kappa(x_i, x_j)$ as defined by the kernel function in Eq. \ref{eq:dist_kernel}.

We differentiate cliques into four types based on the values of the given and predicted labels of the examples. 
The first clique type, denoted by $c_{11}$, is for all pairs of examples $(i, j)$ that share the same given label. If the examples share the same label (i.e., $y_i = y_j$), we assign a negative ``blame'' score (i.e., reward) weighted by $f(c)$ to example $i$ so that it ranks lower in the final rank list of examples sorted by their overall MRF scores. For this clique type, we set the clique weight parameter as $\lambda_c = -1$. 

The second clique type is denoted by $c_{10}$. It is defined for all pairs of examples $(i, j)$ with different labels (i.e., $y_i \neq y_j$) where $y_{j}^\prime = y_{j}$. In this case, example $j$ blames example $i$ for providing an incorrect prediction vote, even though the false vote did not change the prediction output $y_{j}^\prime$ which is consistent with example $j$'s original label $y_j$. For this type, we set $\lambda_c = 1 - \alpha$, where $\alpha$ is a hyper-parameter defined in the range $[0.5,1]$ to control the impact of incorrect vote (i.e., $y_j^\prime \neq y_i)$ on the blame score. 




The third clique type, denoted by $c_{01}$, is for all pairs of examples $(i, j)$ with different labels (i.e., $y_i \neq y_j$) where $y_{j}^\prime \neq y_{j}$ and $y_{j}^\prime \neq y_{i}$. In other words, the prediction output $y_{j}^\prime$ is different from both its own label $y_{j}$ and example $i$'s label $y_i$. While example $i$ did not directly influence the mispredicted label, it did not contribute towards the correct prediction. By setting $\lambda_c = \alpha$, example $j$ assigns a scaled blame score to example $i$.


The fourth clique type, denoted by $c_{00}$, is for all pairs of examples $(i, j)$ with different labels (i.e., $y_i \neq y_j$) where $y_{j}^\prime \neq y_{j}$ and $y_{j}^\prime = y_{i}$. Example $j$ blames example $i$ strongly for supporting a prediction different from its own label $y_j$ which is the same as its prediction $y_j^\prime$. For this type, we set $\lambda_c = \alpha \times b_{f}$ where $b_{f}$ in $[1,inf)$ is the ``blame factor'' and controls the strength of the blame.





\subsubsection{NoiseRank Score Function}


The final ranking function in Eq. \ref{eq:combined_score_func} is the sum of all blame and reward scores accumulated for example $i$ according to Eq. \ref{eq:rankfunction}.

\begin{eqnarray}
\left.
\begin{aligned} 
&P(x_{i},y_{i}|\mathcal{D},Y) \; {\buildrel {rank} \over =}\;          
  \sum_{(x_i, x_j)\in c_{11}} -\kappa(x_i, x_j) \\ 
&+ \sum_{(x_i, x_j) \in c_{10}} (1 - \alpha) \kappa(x_i, x_j) 
+ \sum_{(x_i, x_j) \in c_{01}} \alpha \kappa(x_i, x_j)  \\ 
&+ \sum_{(x_i, x_j) \in c_{00}} (\alpha \times b_f) \kappa(x_i, x_j)
\end{aligned} \right.
\label{eq:combined_score_func}
\end{eqnarray}
The aggregate score $P(x_{i},y_{i}|\mathcal{D},Y)$ is the basis for ranking instances in the dataset. The rank reflects the relative likelihood of being mislabeled and the impact on mispredictions, accounted for in the penalty function. Since the score function is unbounded, detecting label noise given the ranked list requires threshold $\delta \geq 0$ to determine if an instance with detected label noise should be retained ($w=1$) or removed ($w=0$) from the dataset.

\begin{equation} \label{eq:discard}
    w(x_i)=\begin{cases}
    0, & \text{if $P(x_{i},y_{i}|\mathcal{D},Y) > \delta $} \\
    1, & \text{otherwise}
    \end{cases}
\end{equation}




It should be noted that as the dataset size increases, computing the ranking function over all $\mathcal{O}(N|P|)$ pairs is less efficient. Moreover, the value of the feature function $f(c)$ approaches zero as distances between pairs increase. We therefore limit the cliques to the $k$ closest neighbors of example $i$ for assigning blame and reward scores as defined above. This approximation is quite effective especially because the blame and reward scores diminish rapidly and approach zero as distances get larger. 
Another note is that we use the same $k$ value in the score function and in the kernel function (Eq.~\ref{eq:dist_kernel}) for both computing the aggregate rank score function and generating label predictions $y_j^\prime$ for simplification purposes. 






\subsection{Iterative Training}\label{sec:labelnet_classification}
We provide a generic iterative framework to learn classifiers with label noise reduction. As described earlier, the vector space representations of examples in the dataset are used in determining content similarity for noise ranking. Initially, representations can be learned with the available (potentially noisy) labels. In order to improve the learned representations, we can iterate over representation learning, noise ranking and reduction, model training, in order.

The framework is agnostic to the model used for representation learning and classification, depending on the content modality. At a first step, we train the classifier model (eg. standard CNN with simple cross-entropy loss) with the available noisy dataset $D$. The classifier can be used to extract representations for examples in the dataset, which are used to run label noise detection with \algName and remove the examples that are ranked as noisy (i.e., $w(x_i)=0$). Finally, we fine-tune the trained model with the denoised subset of the dataset.

\section{Experiments} 

We report experiments on three public datasets: Food-101N, Clothing1M and YFCC100m on both label noise detection and classification.


\begin{table}[t]
\caption{Dataset Statistics. We use only the noisy (train) labels in NoiseRank. Verified labels (train/validation) are used in other supervised/weakly-supervised methods}
\begin{center} %
\scalebox{0.9}{
\begin{tabular}{ccccc}
\hline
Dataset & \# Classes & \# Train & \# Verified (tr/va) & \# Test \\
\hline 
Food-101N  & 101 & 310K & 55k/5k & 25k
\\
Clothing1M & 14 & 1M & 25k/7k & 10k
\\
YFCC100m & 1000 & 99.2M & -/- & 50k \\
\hline
\end{tabular}
}
\end{center}
\label{tab:datasets}
\end{table}

\textbf{Food-101N} \cite{lee2018cleannet} and \textbf{Clothing-1M} \cite{xiao2015learning}: These are \textbf{real noise} public datasets collected from noisy channels; which are used to study methods for learning in the presence of label noise. These datasets also contain additional verification/clean labels used for noise detection training and validation by supervised label noise reduction methods. Note that for NoiseRank, we do not use these additional verified/clean labels in training or validation. We only use the verified validation labels for evaluation of our method to report results on label noise detection recall and accuracy. These datasets also provide a clean test set with 25K and 10K examples respectively, used for evaluation of the classification task top-1 accuracy. \textbf{YFCC100m} \cite{thomee2015yfcc100m}: This is a large-scale dataset with 99.2M images used in the semi-supervised learning setting in \cite{yalniz2019billion} and we combine it with NoiseRank for detecting label noise in machine-generated labels originated by the semi-supervised learning setup. We use NoiseRank to detect and remove mislabeled examples; and the rest are then leveraged to improve target ImageNet-1k classification. Dataset statistics are summarized in Table~\ref{tab:datasets}. 

\subsection{Experiment setup and hyper-parameters}
For representation learning, we introduced a 256-dimensional bottleneck layer to the ResNet-50 model pre-trained on ImageNet1k. 
First, the pre-trained ResNet-50 is fully fine-tuned with the entire noisy dataset using learning rate 0.002 for [10,10,10] epochs and learning rate decay rate 0.1. 
The output of the bottleneck layer is L2 normalized and used for representing image content. We report results for \algName which conducts label noise detection and removal only once; and iterative \algName wherein after one round of noise removal we fine-tune the ResNet and repeat noise removal. 
For efficient nearest neighbor search, we use open-source library FAISS \cite{JDH17} which takes less than 10 minutes on one GPU for dataset of size 1M with the 256d vector representations. 


\textbf{Unsupervised hyper-parameter selection:} The improvement in data quality can be directly measured (without supervision from verified labels) by the improvement in learnability of the classifier. We measure the training loss at epoch 10 on denoised subsets and select NoiseRank hyper-parameter setting that results in the least training loss. To reduce the parameter search, we first select and fix the best $k$ (number of nearest neighbors) by grid search in $\{5, 10, 20, 50, 100, 250\}$ and then search $\alpha \in \{0.5, 0.6, 0.8\}$ and $b_f \in \{1.0, 1.5, 2.0\}$, and $b=e=1$ in the distance kernel. The ranking cut-off $\delta=0$. 


\subsection{Label Noise Detection Experiments}

\begin{table}[t]
\caption{Label noise detection accuracy. Left: average error rate over all the classes (\%) Right: Label noise recall, F1 and macro-F1 (\%). NoiseRank(I) is iterative NoiseRank}
\begin{center}
\scalebox{0.85}{
\begin{tabular}{lccc}
\hline
& \multicolumn{2}{c}{Average error rate} \\
\hline
Method & Food-101N & Clothing-1M \\
\hline\hline 
\multicolumn{3}{c}{Supervised} \\
\hline
MLP & 10.42 & 16.09 \\
Label Prop \cite{xiaojin2002learning} & 13.24 & 17.81 \\
Label Spread \cite{zhou2004learning} & 12.03 & 17.71 \\
CleanNet \cite{lee2018cleannet} & 6.99 & 15.77 \\
\hline
\multicolumn{3}{c}{Weakly-Supervised} \\
\hline
Cls. Filt. & 16.60 & 23.55 \\
Avg. Base. \cite{lee2018cleannet} & 16.20 & 30.56 \\
\hline
\multicolumn{3}{c}{Unsupervised} \\
\hline
DRAE \cite{xia2015learning} & 18.70 & 38.95 \\
unsup-kNN & 26.63 & 43.31 \\
\algName & 24.02 & 23.54 \\
\textbf{\algName (I)} & \textbf{18.43} & \textbf{22.81} \\
\hline
\end{tabular}
}
\quad
\scalebox{0.9}{
\renewcommand{\arraystretch}{1}
\begin{tabular}{llcccc}
\hline
Method & Type & Recall & F1 & MacroF1  \\
\hline\hline 
\multicolumn{5}{c}{Food-101N (19.66\% estimated noise)} \\
\hline
CleanNet & sup. & 71.06 & \textbf{74.01} & \textbf{84.04}  \\
Avg. Base. & weakly & 47.70 & 59.57 & 76.08 \\
unsup-kNN & unsup. & 22.02 & 24.23 & 54.03 \\
 \algName (I) & unsup. & \textbf{85.61} & 64.42 & 76.06  \\
\hline
\multicolumn{5}{c}{Clothing-1M (38.46\% estimated noise)} \\
\hline
CleanNet & sup. & 69.40 & \textbf{73.99} & \textbf{79.65} \\
Avg. Base. & weakly & 43.92 & 55.14 & 67.65  \\
unsup-kNN & unsup. & 10.85 & 16.60 & 44.26  \\
\algName (I) & unsup. & \textbf{74.18} & 71.74 & 76.52  \\
\hline
\end{tabular}
}
\end{center}
\label{tab:detection}
\end{table}

We report the effectiveness of our proposed method on detecting label noise in Table~\ref{tab:detection}, in terms of i) averaged detection error rate over all classes in Food101-N and Clothing-1M, and ii) in terms of label noise recall and F1. Table \ref{tab:detection} details the average error rate of label noise detection on the verified validation set compared against a wide range of baselines, as reported in \cite{lee2018cleannet}. The naive baseline predicts all samples as correctly labeled, and therefore its error rate approximates the true noise distribution assuming a random selection of the ground truth set. Clothing-1M has a significant amount of noise estimated at 38.46\%. In this significantly noisy dataset, iterative NoiseRank even as an unsupervised method, strongly outperforms unsupervised outlier removal method DRAE \cite{xia2015learning} by a large margin of 16.15\% (which is 40\% error reduction) and weakly supervised Average Baseline (Avg. Base.) \cite{lee2018cleannet} by 7.75\% (which is 25\% error reduction) on avg error rate. This is state-of-the-art noise detection error rate among unsupervised alternatives on this dataset. Avg. Base. computes the cosine similarity between an instance representation and the averaged representation of a class; and although it does not use verified labels in training, it uses them to select the threshold on cosine similarity for label noise detection. On Food101-N the estimated noise is 19.66\% and the avg error rate of iterative NoiseRank and DRAE are comparable.

However, since noise vs. clean instance distribution is imbalanced, we further measure recall, F1 and macro-F1 scores for label noise detection in Table \ref{tab:detection}. NoiseRank has state-of-the-art recall of 85.61\% on Food-101N and 74.18\% on Clothing-1M. NoiseRank F1/MacroF1 is competitive with the best supervised method in noise detection CleanNet \cite{lee2018cleannet} which requires verified labels in training and validation, and thus has a significant advantage compared to unsupervised and weakly-supervised methods. It should be noted that effective noise recall directly impacts  classification, and is therefore an important evaluation metric for label noise detection and removal.

\begin{table}[t]
\caption{Image classification on Food-101N results in terms of top-1 accuracy (\%). Train data (310k) and test data (25k). CleanNet is trained with an additional 55k/5k (tr/va) verification labels to provide the required supervision on noise detection}
\begin{center}
\scalebox{0.9}{
\begin{tabular}{lcccc}
\hline
\textbf{\#} & Method & Training & Pre-training & Top-1 \\
\hline\hline 
1 & None \cite{lee2018cleannet} & noisy train & ImageNet & 81.44 \\
\hline
2 & CleanNet \cite{lee2018cleannet} & noisy(+verified) & ImageNet & 83.95 \\
\hline
3 & DeepSelf \cite{han2019deep} & noisy train & ImageNet & 85.11 \\
\hline
4 & \algName & cleaned train & ImageNet & 85.20 \\ 
5 &  & cleaned train & noisy train \#1 & \textbf{85.78} \\ 
\hline
\end{tabular}
}
\end{center}
\label{tab:classification_food}
\end{table}

\subsection{Classification Experiments}
We conducted experiments to study the impact of data quality on the classification task using the ResNet-50 classifier pretrained on ImageNet and initially fine-tuned on noisy dataset and later on denoised subset with NoiseRank.
In results table \ref{tab:classification_food} and \ref{tab:classification_clothing}, in each row, the model is fine-tuned with the mentioned training examples on the specified pre-trained model (eg. ``noisy train \# 1" refers to the model \# 1 referenced in the table that was trained using noisy training samples on ImageNet pre-trained Resnet-50). Similarly, in Table~\ref{tab:classification_clothing}, ``\# 4" in the pre-training column, refers to model \# 4 indicated in the table.


In Table \ref{tab:classification_food} Food101-N, NoiseRank achieves state of the art 85.78\% in top-1 accuracy compared to unsupervised \cite{han2019deep}'s 85.11\%, and 11\% error reduction over supervised noise reduction method CleanNet. This can be attributed to the high noise recall on Food-101N as examined earlier. In Table \ref{tab:classification_clothing} Clothing-1M, NoiseRank used to reduce label noise in noisy train ($\sim$40\% estimated noise) is effective in improving classification from 68.94\% to 73.82\% (16\% error reduction), even without supervision from clean set in high noise regime, and performs comparable to recent unsupervised \cite{han2019deep} and marginally outperforms unsupervised PENCIL \cite{yi2019prob}. In contrast to \cite{han2019deep} and \cite{yi2019prob}, NoiseRank framework allows for flexible choice of classifier and optimization/loss function, and yet achieves comparable improvement due to noise reduction, using standard cross-entropy loss and standard training framework. This underlines the benefits of the proposed framework without compromising on classification improvements. Supervised baselines CleanNet and Loss Correction respectively utilize additional verified labels, and yet the performance gain from noise removal for ours is highly competitive, even in this high noise regime. Lastly, we also reported results of fine-tuning each method with an additional clean 50k set, as per the setting followed in \cite{patrini2017making}. \cite{han2019deep} achieves best result of 81.16\% with clean 50k sample set. We note that even without noise correction, the inclusion of the clean set boosts accuracy from 68.94\% to 79.43\% and may shadow the benefit of noise removal; with CleanNet \cite{lee2018cleannet}  at 79.90\% and ours at 79.57\% being comparable in this setting.



\begin{table}[t]
\caption{Image classification on Clothing-1M results in terms of top-1 accuracy (\%). Train data (1M) and test data (10k). CleanNet and Loss Correction are trained with an additional 25k/7k (train/validation) verification labels to provide required supervision on noise detection/correction
}
\begin{center}
\scalebox{0.9}{
\begin{tabular}{lcccc}
\hline
\# & Method & Training & Pre-training & Top-1 \\
\hline\hline 
1 & None \cite{patrini2017making} & clean50k & ImageNet & 75.19 \\
\hline
2 & None \cite{patrini2017making} & noisy train & ImageNet & 68.94 \\
3 &  & clean50k & noisy train \# 2 & 79.43 \\
\hline
4 & loss cor.\cite{patrini2017making} & noisy(+verified) & ImageNet & 69.84 \\
5 &  & clean50k & \# 4 & 80.38 \\
\hline
6 & Joint opt. \cite{tanaka2018joint} & noisy train & ImageNet & 72.16 \\
\hline
7 & PENCIL \cite{yi2019prob} & noisy train & ImageNet & 73.49 \\
\hline
8 & CleanNet \cite{lee2018cleannet} & noisy(+verified) & ImageNet & 74.69 \\
9 &  & clean50k & \# 8 & 79.90 \\
\hline
10 & DeepSelf \cite{han2019deep} & noisy train & ImageNet & 74.45 \\
11 &  & clean50k & \# 10 & \textbf{81.16} \\
\hline
12 &  & cleaned train & ImageNet & 73.77 \\
13 & \algName & cleaned train & noisy train \#2 & 73.82 \\
14 &  & clean50k & \# 13 & 79.57 \\
\hline
\end{tabular}
}
\end{center}
\label{tab:classification_clothing}
\end{table}



\begin{table}[ht]
\centering
\caption{Left: ImageNet benchmark top-1 accuracy (\%). \algName with removal of top x\% ranked instances in noisy machine generated labels, against random removal. Right: Examples of noisy machine generated labels detected by NoiseRank (M: mislabeled instances, C: correctly labeled instances mistakenly identified by NoiseRank)}
\begin{tabular}{cc}
\label{table:classification_flickr}
    \scalebox{0.85}{
    \begin{tabular}{lc}
    \hline
    Method & Top-1 Accuracy \\
    \hline\hline 
    None \cite{yalniz2019billion} & 79.06 \\
    \hline
    Top 0.6\% removed & 79.13 \\
    Top 1.2\% removed & 79.12 \\
    Top 1.8\% removed & \textbf{79.34} \\
    \hline
    Random 1.8\% removed & 78.96 \\
    \hline
    \end{tabular}
    }
    &~~
    \includegraphics[width=0.5\linewidth,height=3cm,valign=m]{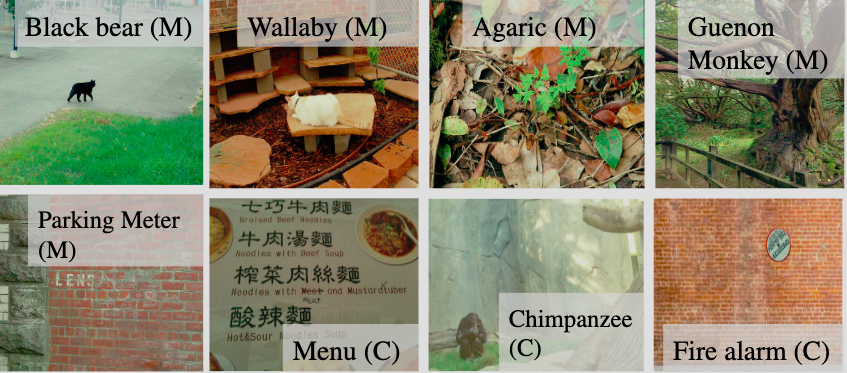}
    \end{tabular}
\end{table}

In Table \ref{table:classification_flickr}, we report semi-supervised learning results on large YFCC100m dataset, with and without label denoising. \cite{yalniz2019billion} is used to train a ResNet-101 to label images in YFCC100m into 1K ImageNet classes. 16K images from each class that have the most confident machine label predictions are retained. However, even after filtering, these labels contain noise as shown by examples detected using NoiseRank (right: Table~\ref{table:classification_flickr}). We run NoiseRank to remove label noise from the 16M machine labeled images. The denoised images are used to pre-train ResNet-50, then fine-tuned with ImageNet-1K train set and evaluated on benchmark ImageNet-1K test set. The top-1 accuracy without noise removal is 79.06\% and with noise removal is 79.34\%, in comparison to removing the same number of random instances (78.96\%) averaged over three runs. Note that in this setting, noise removal is not applied directly to the target classification task, but rather to the dataset used to pre-train the model, later trained on the target dataset.



\subsection{Interpretability Analysis}

\begin{figure*}[t]
\begin{center}
\begin{subfigure}[t]{0.45\textwidth}
\includegraphics[width=5.5cm,height=3cm]{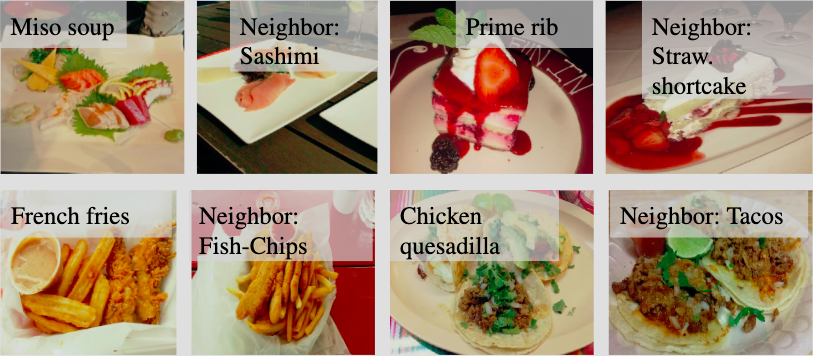}
\caption{Examples correctly detected as label noise in Food-101N shown along with its nearest neighbor}
\label{fig:food_correct}
\end{subfigure}
~~
\begin{subfigure}[t]{0.45\textwidth}
\includegraphics[width=5.5cm,height=3cm]{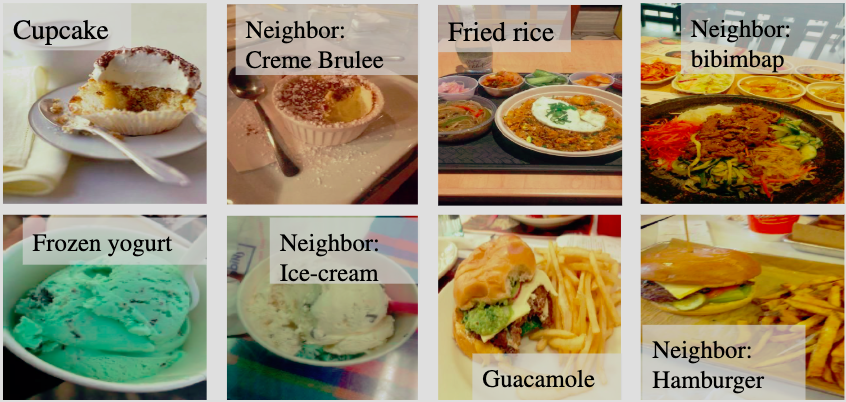}
\caption{Examples incorrectly identified as label noise by NoiseRank, shown along with their nearest neighbors (top row); Examples identified as label noise by NoiseRank but are wrongly verified by humans (bottom row) in Food101-N}
\label{fig:food_incorrect}
\end{subfigure}
\end{center}
\caption{Interpretability analysis of NoiseRank predictions on Food101-N}
\end{figure*}

\begin{figure*}[t]
\begin{center}
\begin{subfigure}[t]{0.3\textwidth}
\includegraphics[width=4cm,height=3cm]{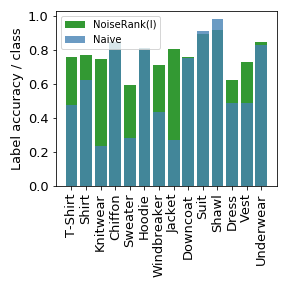}
\caption{Accuracy per class for label noise detection}
\label{fig:clothing_perclassacc}
\end{subfigure}
~~
\begin{subfigure}[t]{0.3\textwidth}
\includegraphics[width=4cm,height=4.5cm]{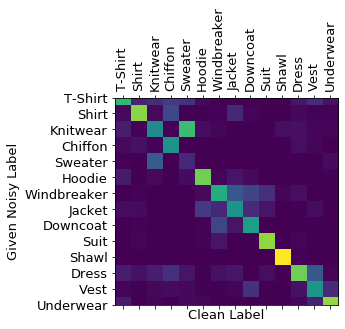}
\caption{Noise transition matrix between classes, estimated on validation set}
\label{fig:closthing_noisetransition}
\end{subfigure}
~~
\begin{subfigure}[t]{0.3\textwidth}
\includegraphics[width=4cm,height=4.5cm]{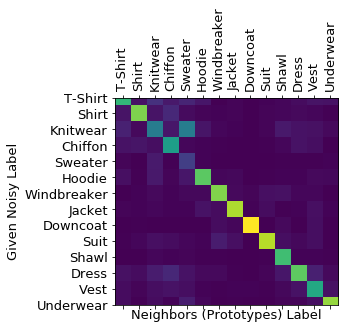}
\caption{NoiseRank score function matrix between class types}
\label{fig:clothing_blame_matrix}
\end{subfigure}
\end{center}
\caption{Interpretability analysis of NoiseRank predictions on Clothing-1M}
\label{fig:clothing_class_analysis}
\end{figure*}

Interpretability is useful in providing explanations about the predictions made by machine learning algorithms. NoiseRank is a transparent framework that can be easily used to provide human level analysis of why a given example was predicted as mislabeled or not mislabeled.

In Fig.~\ref{fig:food_correct}, we show sample images correctly identified by NoiseRank as label noise in the Food-101N dataset, along with their nearest neighbors. The nearest neighbors provide supporting visual evidence towards understanding the prediction made by NoiseRank. Interpretability is also useful for identifying hard instances; to support building better datasets and models. In Fig.~\ref{fig:food_incorrect} (top row) we show sample images incorrectly identified as label noise in Food-101N. As seen, these are tough examples with contradictory labels to their nearest neighbors, which provides insights into when and which instances might be confused with others. The bottom row shows sample images identified as label noise by NoiseRank but verified (seemingly incorrectly) by humans. Such samples further justify our belief that human label verification can also be prone to errors.

In Fig.~\ref{fig:clothing_class_analysis}, we provide class-wise analysis on the 14 Clothing-1M dataset class types. Fig.~\ref{fig:clothing_perclassacc} reports noise detection accuracy per class (NoiseRank), compared to the original noise ratio in the dataset (Naive). Fig~\ref{fig:closthing_noisetransition} provides the estimated probability of flipping a clean label of one class to another (noise transition matrix) estimated on the validation set. In Fig.~\ref{fig:clothing_blame_matrix}, we visualize how NoiseRank scores each example based on its neighboring  prototypes. Each cell in ~\ref{fig:clothing_blame_matrix} maps a given noisy label class and its prototype neighbors' class aggregated over the pairs used in the scoring function (Eq.~\ref{eq:combined_score_func}), with weight equal to its contribution in the score function; and the matrix is then column-normalized for the distribution. It implicitly encodes the noise transition probability between class types without any knowledge of the clean labels, as seen from its similarity to \ref{fig:closthing_noisetransition}. 



\section{Conclusion}
In this paper, we proposed an unsupervised label noise ranking algorithm based on Markov Random Fields. The dependence model is used to estimate and rank instances by the posterior probability of being incorrectly labeled in the dataset. We evaluated the framework on real noise datasets and showed its effectiveness on improving image classification, even compared to supervised alternatives. The proposed iterative framework has advantages over recent unsupervised methods in that it provides interpretable label noise detection and ranking, and is agnostic to classifier architecture and optimization framework (works for any content modality with standard widely used deep learning models without constraining the choice of model, loss function, or optimizer).


\clearpage

\bibliographystyle{splncs04}
\bibliography{egbib}

\end{document}